\documentclass{llncs}
\usepackage{makeidx}  
\usepackage{url}      
\usepackage{subfigure}
\usepackage{amsfonts}
\usepackage{amsmath}
\usepackage{cancel}
\usepackage{graphicx}
\usepackage{color}
\usepackage{hyperref}

\usepackage{graphicx}
\graphicspath{{pics/}{figs/}}

\begin{document}
\frontmatter          
\pagestyle{headings}  
\mainmatter              
\title{V-Net: Fully Convolutional Neural Networks for Volumetric Medical Image Segmentation}
%
\titlerunning{V-Net: Fully Convolutional Neural Networks for Volumetric Medical Image Segmentation}  


\author{Fausto Milletari \inst{1},
	Nassir Navab\inst{1,2},
	Seyed-Ahmad Ahmadi \inst{3}
}

\institute{
  Computer Aided Medical Procedures, Technische Universit\"{a}t M\"{u}nchen, Germany 
  \and
  Computer Aided Medical Procedures,  Johns Hopkins University, Baltimore, USA 
  \and
  Department of Neurology, Klinikum Grosshadern, Ludwig-Maximilians-Universit\"{a}t M\"{u}nchen, Germany}

\authorrunning{F. Milletari \and A. Ahmadi  \and N. Navab}   

\maketitle              

\begin{abstract}
Convolutional Neural Networks (CNNs) have been recently employed to solve problems from both the computer vision and medical image analysis fields. Despite their popularity, most approaches are only able to process 2D images while most medical data used in clinical practice consists of 3D volumes. In this work we propose an approach to 3D image segmentation based on a volumetric, fully convolutional, neural network. Our CNN is trained end-to-end on MRI volumes depicting prostate, and learns to predict segmentation for the whole volume at once. We introduce a novel objective function, that we optimise during training, based on Dice coefficient. In this way we can deal with situations where there is a strong imbalance between the number of foreground and background voxels. To cope with the limited number of annotated volumes available for training, we augment the data applying random non-linear transformations and histogram matching. We show in our experimental evaluation that our approach achieves good performances on challenging test data while requiring only a fraction of the processing time needed by other previous methods.
\end{abstract}

\section{Introduction and Related Work}
\label{sec:intro}
Recent research in computer vision and pattern recognition has highlighted the capabilities of Convolutional Neural Networks (CNNs) to solve challenging tasks such as classification, segmentation and object detection, achieving state-of-the-art performances. 
This success has been attributed to the ability of CNNs to learn a hierarchical representation of raw input data, without relying on handcrafted features. 
As the inputs are processed through the network layers, the level of abstraction of the resulting features increases. 
Shallower layers grasp local information while deeper layers use filters whose receptive fields are much broader that therefore capture global information \cite{zeiler2014visualizing}. 

\begin{figure} 	
\centering 	
\includegraphics[scale=0.17]{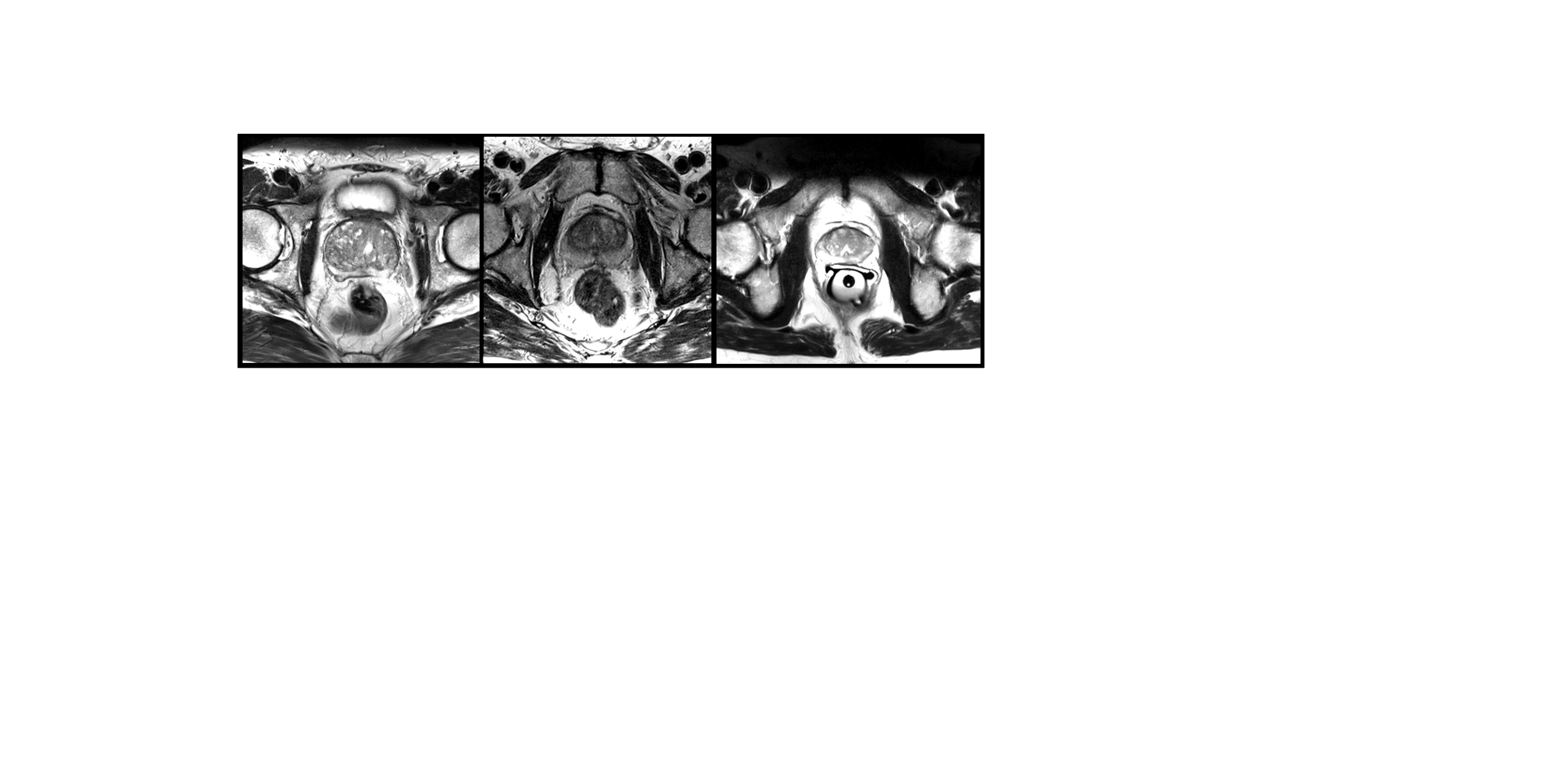} 	
\caption{Slices from MRI volumes depicting prostate. This data is part of the PROMISE2012 challenge dataset \cite{litjens2014evaluation}.} \label{fig:anatomies} 
\end{figure}

Segmentation is a highly relevant task in medical image analysis. 
Automatic delineation of organs and structures of interest is often necessary to perform tasks such as visual augmentation \cite{moradi2009augmenting}, computer assisted diagnosis \cite{porter2003combining}, interventions \cite{zettinig2015multimodal} and extraction of quantitative indices from images \cite{bernard2015standardized}. 
In particular, since diagnostic and interventional imagery often consists of 3D images, being able to perform volumetric segmentations by taking into account the whole volume content at once, has a particular relevance. 
In this work, we aim to segment prostate MRI volumes. This is a challenging task due to the wide range of appearance the prostate can assume in different scans due to deformations and variations of the intensity distribution. Moreover, MRI volumes are often affected by artefacts and distortions due to field inhomogeneity. Prostate segmentation is nevertheless an important task having clinical relevance both during diagnosis, where the volume of the prostate needs to be assessed \cite{roehrborn1999serum}, and during treatment planning, where the estimate of the anatomical boundary needs to be accurate \cite{huyskens2009qualitative,zettinig2015multimodal}. 

CNNs have been recently used for medical image segmentation. 
Early approaches obtain anatomy delineation in images or volumes by performing patch-wise image classification. Such segmentations are obtained by only considering local context and therefore are prone to failure, especially in challenging modalities such as ultrasound, where a high number of mis-classified voxel are to be expected.
Post-processing approaches such as connected components analysis normally yield no improvement and therefore, more recent works, propose to use the network predictions in combination with Markov random fields  \cite{kamnitsas2016efficient}, voting strategies \cite{milletari2016hough} or more traditional approaches such as level-sets  \cite{cha2016urinary}. 
Patch-wise approaches also suffer from efficiency issues. When densely extracted patches are processed in a CNN, a high number of computations is redundant and therefore the total algorithm runtime is high. In this case, more efficient computational schemes can be adopted.

Fully convolutional network trained end-to-end were so far applied only to 2D images both in computer vision \cite{noh2015learning,long2015fully} and microscopy image analysis \cite{ronneberger2015u}. These models, which served as an inspiration for our work, employed different network architectures and were trained to predict a segmentation mask, delineating the structures of interest, for the whole image. In \cite{noh2015learning} a pre-trained VGG network architecture \cite{simonyan2014very} was used in conjunction with its mirrored, de-convolutional, equivalent to segment RGB images by leveraging the descriptive power of the features extracted by the innermost layer. In \cite{long2015fully} three fully convolutional deep neural networks, pre-trained on a classification task, were refined to produce segmentations while in \cite{ronneberger2015u} a brand new CNN model, especially tailored to tackle biomedical image analysis problems in 2D, was proposed.

In this work we present our approach to medical image segmentation that leverages the power of a fully convolutional neural networks, trained end-to-end, to process MRI volumes. 
Differently from other recent approaches we refrain from processing the input volumes slice-wise and we propose to use volumetric convolutions instead. We propose a novel objective function based on Dice coefficient maximisation, that we optimise during training.
We demonstrate fast and accurate results on prostate MRI test volumes and we provide direct comparison with other methods which were evaluated on the same test data \footnote{Detailed results available on \url{http://promise12.grand-challenge.org/results/}}. 

\section{Method}
\label{sec:method}

\begin{figure} 	
\centering 	
\includegraphics[scale=0.26]{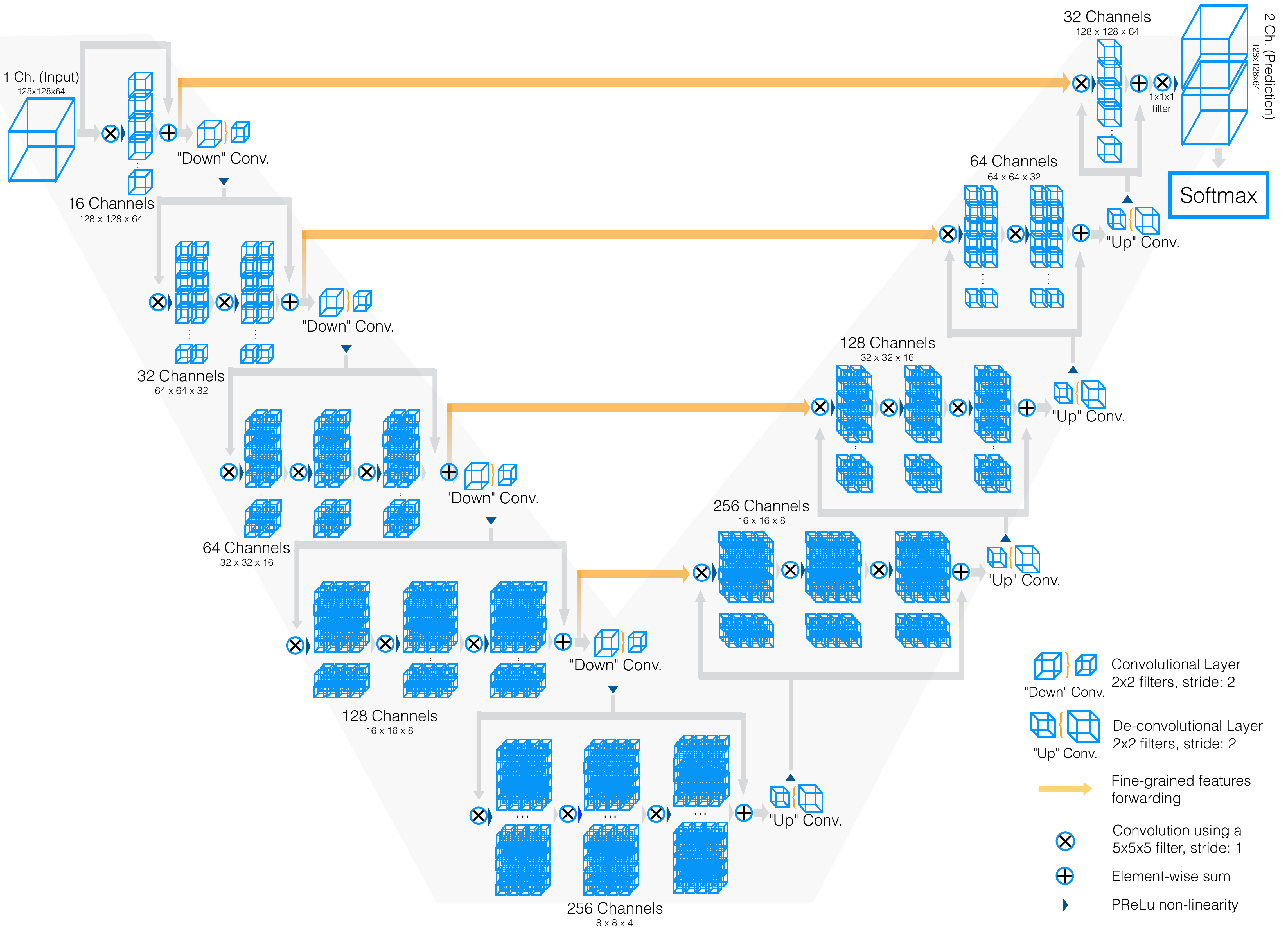} 	
\caption{Schematic representation of our network architecture. Our custom implementation of Caffe \cite{jia2014caffe} processes 3D data by performing volumetric convolutions. Best viewed in electronic format.} \label{fig:VnetImage} 
\end{figure}

In Figure \ref{fig:VnetImage} we provide a schematic representation of our convolutional neural network. 
We perform convolutions aiming to both extract features from the data and, at the end of each stage, to reduce its resolution by using appropriate stride. The left part of the network consists of a compression path, while the right part decompresses the signal until its original size is reached. Convolutions are all applied with appropriate padding.

The left side of the network is divided in different stages that operate at different resolutions. Each stage comprises one to three convolutional layers. Similarly to the approach presented in \cite{he2015deep}, we formulate each stage such that it learns a residual function: the input of each stage is (a) used in the convolutional layers and processed through the non-linearities and (b) added to the output of the last convolutional layer of that stage in order to enable learning a residual function. As confirmed by our empirical observations, this architecture ensures convergence in a fraction of the time required by a similar network that does not learn residual functions. 

\begin{figure} 	
\centering 	
\includegraphics[scale=0.34]{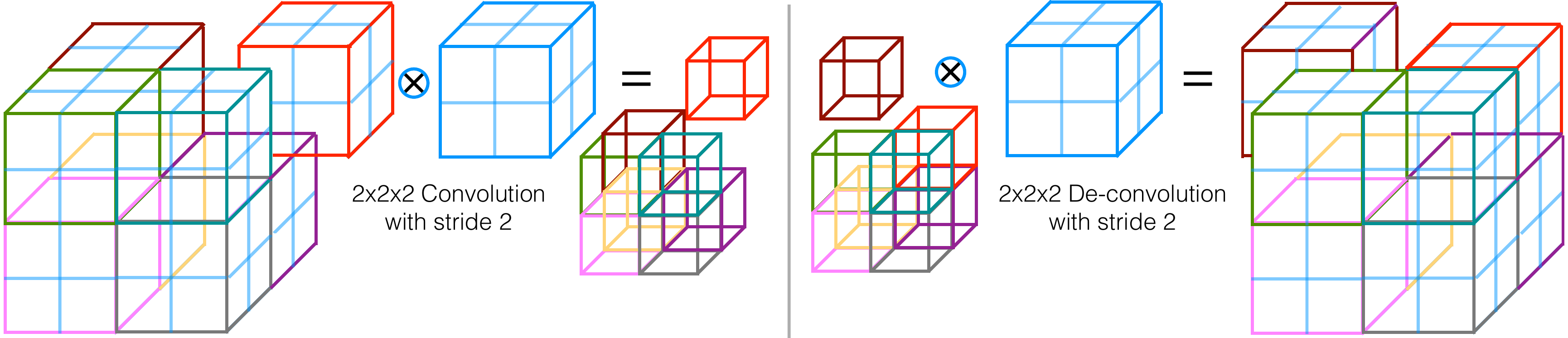} 	
\caption{Convolutions with appropriate stride can be used to reduce the size of the data. Conversely, de-convolutions increase the data size by projecting each input voxel to a bigger region through the kernel.} \label{fig:updown} 
\end{figure}

The convolutions performed in each stage use volumetric kernels having size $5\times5\times5$ voxels.
As the data proceeds through different stages along the compression path, its resolution is reduced. This is performed through convolution with $2\times2\times2$ voxels wide kernels applied with stride $2$ (Figure \ref{fig:updown}). Since the second operation extracts features by considering only non overlapping $2\times2\times2$ volume patches, the size of the resulting feature maps is halved. 
This strategy serves a similar purpose as pooling layers that, motivated by \cite{springenberg2014striving} and other works discouraging the use of max-pooling operations in CNNs, have been replaced in our approach by convolutional ones. Moreover, since the number of feature channels doubles at each stage of the compression path of the V-Net, and due to the formulation of the model as a residual network, we resort to these convolution operations to double the number of feature maps as we reduce their resolution. PReLu non linearities are applied throughout the network.

Replacing pooling operations with convolutional ones results also to networks that, depending on the specific implementation, can have a smaller memory footprint during training, due to the fact that no switches mapping the output of pooling layers back to their inputs are needed for back-propagation, and that can be better understood and analysed \cite{zeiler2014visualizing} by applying only de-convolutions instead of un-pooling operations. 

Downsampling allows us to reduce the size of the signal presented as input and to increase the receptive field of the features being computed in subsequent network layers. Each of the stages of the left part of the network, computes a number of features which is two times higher than the one of the previous layer.

The right portion of the network extracts features and expands the spatial support of the lower resolution feature maps in order to gather and assemble the necessary information to output a two channel volumetric segmentation. The two features maps computed by the very last convolutional layer, having $1\times1\times1$ kernel size and producing outputs of the same size as the input volume, are converted to probabilistic segmentations of the foreground and background regions by applying soft-max voxelwise.
After each stage of the right portion of the CNN, a de-convolution operation is employed in order increase the size of the inputs (Figure \ref{fig:updown}) followed by one to three convolutional layers involving half the number of $5\times5\times5$ kernels employed in the previous layer. Similar to the left part of the network, also in this case we resort to learn residual functions in the convolutional stages.

Similarly to \cite{ronneberger2015u}, we forward the features extracted from early stages of the left part of the CNN to the right part. This is schematically represented in Figure \ref{fig:VnetImage} by horizontal connections. In this way we gather fine grained detail that would be otherwise lost in the compression path and we improve the quality of the final contour prediction. We also observed that when these connections improve the convergence time of the model.

We report in Table \ref{table:receptiveFields} the receptive fields of each network layer, showing the fact that the innermost portion of our CNN already captures the content of the whole input volume. We believe that this characteristic is important during segmentation of poorly visible anatomy: the features computed in the deepest layer perceive the whole anatomy of interest at once, since they are computed from data having a spatial support much larger than the typical size of the anatomy we seek to delineate, and therefore impose global constraints.

\begin{table}
\begin{centering}
\protect\caption{Theoretical receptive field of the $3\times3\times3$ convolutional layers of the network.} 
\begin{tabular}{|c|c|c||c|c|c|}
\hline 
Layer & Input Size & Receptive Field & Layer & Input Size & Receptive Field\tabularnewline
\hline 
\hline 
L-Stage 1 & $128$ & $5\times5\times5$ & R-Stage 4 & 16 & $476\times476\times476$\tabularnewline
\hline 
L-Stage 2 & $64$ & $22\times22\times22$ & R-Stage 3 & 32 & $528\times528\times528$\tabularnewline
\hline 
L-Stage 3 & $32$ & $72\times72\times72$ & R-Stage 2 & 64 & $546\times546\times546$\tabularnewline
\hline 
L-Stage 4 & $16$ & $172\times172\times172$ & R-Stage 1 & 128 & $551\times551\times551$\tabularnewline
\hline 
L-Stage 5 & $8$ & $372\times372\times372$ & Output & 128 & \textbf{$551\times551\times551$}\tabularnewline
\hline 
\end{tabular} \label{table:receptiveFields}
\par\end{centering}
\end{table}

\section{Dice loss layer}
\begin{figure} 	
\centering 	
\includegraphics[scale=0.208]{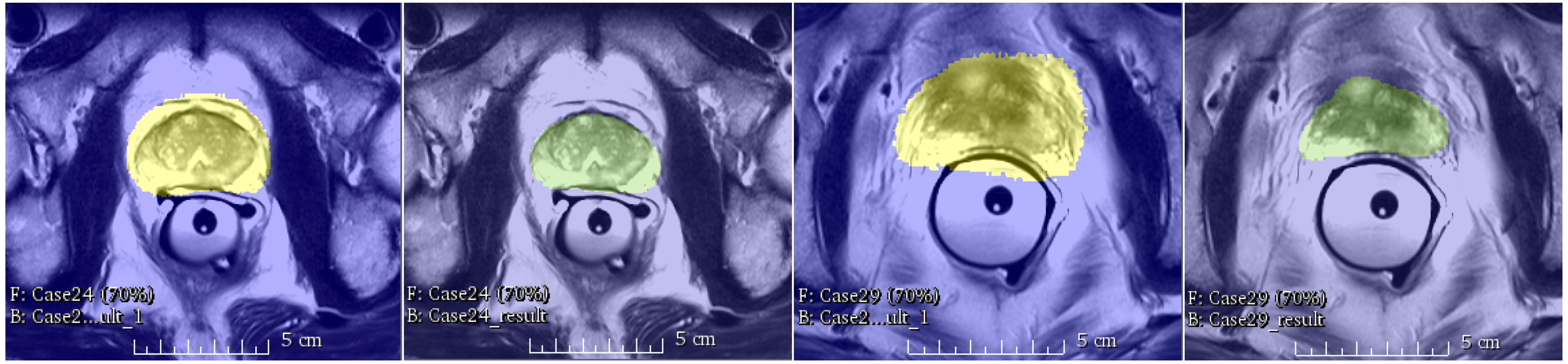} 	
\caption{Qualitative comparison between the results obtained using the Dice coefficient based loss (green) and re-weighted soft-max with loss (yellow).} \label{fig:qualitativecomparison} 
\end{figure}

The network predictions, which consist of two volumes having the same resolution as the original input data, are processed through a soft-max layer which outputs the probability of each voxel to belong to foreground and to background. In medical volumes such as the ones we are processing in this work, it is not uncommon that the anatomy of interest occupies only a very small region of the scan. This often causes the learning process to get trapped in local minima of the loss function yielding a network whose predictions are strongly biased towards background. As a result the foreground region is often missing or only partially detected. Several previous approaches resorted to loss functions based on sample re-weighting where foreground regions are given more importance than background ones during learning. In this work we propose a novel objective function based on dice coefficient, which is a quantity ranging between $0$ and $1$ which we aim to maximise. The dice coefficient $D$ between two binary volumes can be written as
\[
D=\frac{2\sum_{i}^{N}p_{i}g_{i}}{\sum_{i}^{N}p_{i}^{2}+\sum_{i}^{N}g_{i}^{2}}
\]

where the sums run over the $N$ voxels, of the predicted binary segmentation volume $p_i\in{P}$ and the ground truth binary volume $g_i\in{G}$. This formulation of Dice can be differentiated yielding the gradient  
\[
\frac{\partial D}{\partial p_{j}}=2\left[\frac{g_{j}\left(\sum_{i}^{N}p_{i}^{2}+\sum_{i}^{N}g_{i}^{2}\right)-2p_{j}\left(\sum_{i}^{N}p_{i}g_{i}\right)}{\left(\sum_{i}^{N}p_{i}^{2}+\sum_{i}^{N}g_{i}^{2}\right)^{2}}\right]
\]
computed with respect to the $j$-th voxel of the prediction. Using this formulation we do not need to assign weights to samples of different classes to establish the right balance between foreground and background voxels, and we obtain results that we experimentally observed are much better than the ones computed through the same network trained optimising a multinomial logistic loss with sample re-weighting (Fig. \ref{fig:qualitativecomparison}). 

\subsection{Training}
Our CNN is trained end-to-end on a dataset of prostate scans in MRI. An example of the typical content of such volumes is shown in Figure \ref{fig:anatomies}. All the volumes processed by the network have fixed size of $128\times128\times64$ voxels and a spatial resolution of $1\times1\times1.5$ millimeters.

Annotated medical volumes are not easy to obtain due to the fact that one or more experts are required to manually trace a reliable ground truth annotation and that there is a cost associated with their acquisition. In this work we found necessary to augment the original training dataset in order to obtain robustness and increased precision on the test dataset. 

During every training iteration, we fed as input to the network randomly deformed versions of the training images by using a dense deformation field obtained through a $2\times2\times2$ grid of control-points and B-spline interpolation. This augmentation has been performed "on-the-fly", prior to each optimisation iteration, in order to alleviate the otherwise excessive storage requirements. Additionally we vary the intensity distribution of the data by adapting, using histogram matching, the intensity distributions of the training volumes used in each iteration, to the ones of other randomly chosen scans belonging to the dataset.

\subsection{Testing}
A Previously unseen MRI volume can be segmented by processing it in a feed-forward manner through the network. The output of the last convolutional layer, after soft-max, consists of a probability map for background and foreground. The voxels having higher probability ($>0.5$) to belong to the foreground than to the background are considered part of the anatomy.

\section{Results}
\label{sec:results}

\begin{figure} 	
\centering 	
\includegraphics[scale=0.18]{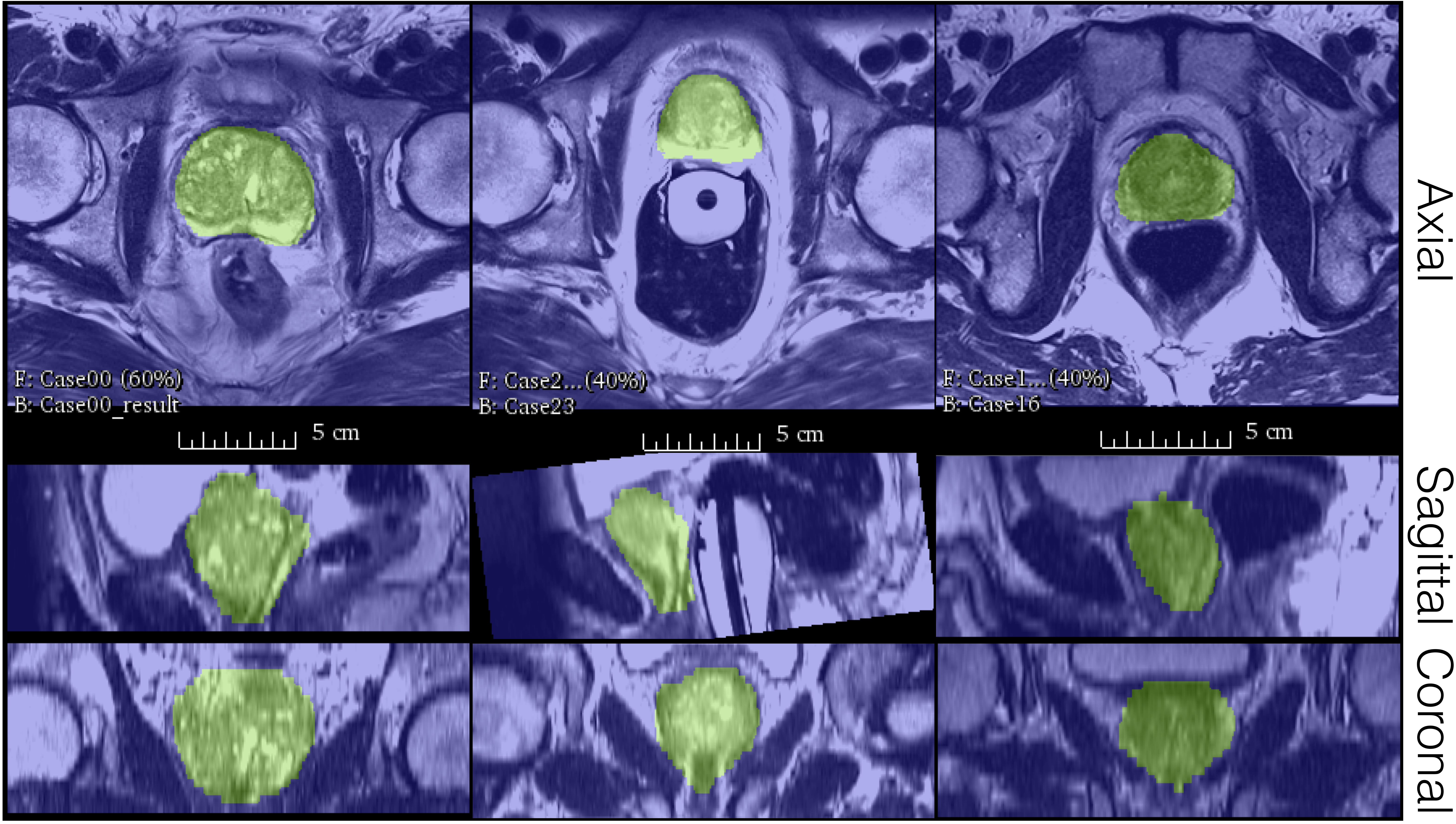} 	
\caption{Qualitative results on the PROMISE 2012 dataset \cite{litjens2014evaluation}.} \label{fig:qualitative} 
\end{figure}

We trained our method on $50$ MRI volumes, and the relative manual ground truth annotation, obtained from the "PROMISE2012" challenge dataset \cite{litjens2014evaluation}. This dataset contains medical data acquired in different hospitals, using different equipment and different acquisition protocols. The data in this dataset is representative of the clinical variability and challenges encountered in clinical settings. As previously stated we massively augmented this dataset through random transformation performed in each training iteration, for each mini-batch fed to the network. The mini-batches used in our implementation contained two volumes each, mainly due to the high memory requirement of the model during training. We used a momentum of $0.99$ and a initial learning rate of $0.0001$ which decreases by one order of magnitude every $25$K iterations. 

\begin{figure} 	
\centering 	
\includegraphics[scale=0.19]{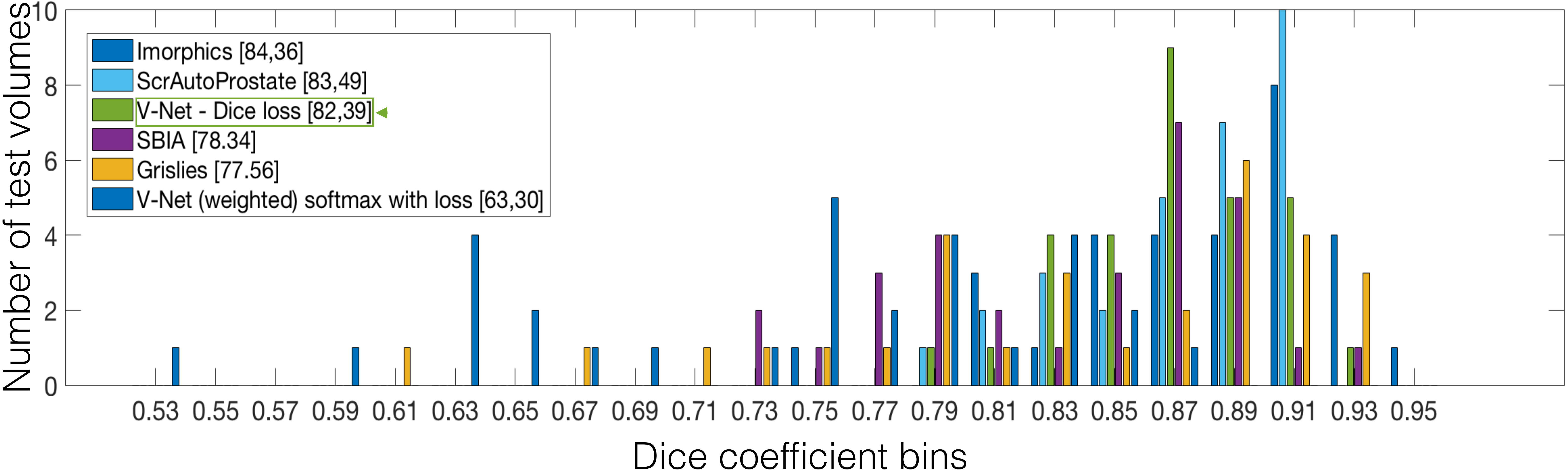} 	
\caption{Distribution of volumes with respect to the Dice coefficient achieved during segmentation.} 
\label{fig:hist} 
\end{figure}

We tested V-Net on $30$ MRI volumes depicting prostate whose ground truth annotation was secret. All the results reported in this section of the paper were obtained directly from the organisers of the challenge after submitting the segmentation obtained through our approach. The test set was representative of the clinical variability encountered in prostate scans in real clinical settings \cite{litjens2014evaluation}. 

We evaluated the approach performance in terms of Dice coefficient, Hausdorff distance of the predicted delineation to the ground truth annotation and in terms of score obtained on the challenge data as computed by the organisers of "PROMISE 2012" \cite{litjens2014evaluation}. The results are shown in Table \ref{tab:res} and Fig. \ref{fig:hist}.

\begin{table}
\caption{Quantitative comparison between the proposed approach and the current best results on the PROMISE 2012 challenge dataset.} \label{tab:res}
\begin{tabular}{|c|c|c|c|}
\hline 
Algorithm & Avg. Dice & Avg. Hausdorff distance & Score on challenge task\tabularnewline
\hline 
\hline 
V-Net + Dice-based loss & $0.869 \pm 0.033$ & $5.71 \pm 1.20$ mm & $82.39$\tabularnewline
\hline 
V-Net + mult. logistic loss & $0.739 \pm 0.088 $ & $10.55 \pm 5.38$ mm & $63.30$\tabularnewline
\hline 
Imorphics \cite{imorp} & $0.879 \pm 0.044$ & $5.935 \pm 2.14 $ mm & $84.36$\tabularnewline
\hline 
ScrAutoProstate & $0.874 \pm 0.036$ & $5.58 \pm 1.49 $ mm & $83.49$ \tabularnewline
\hline
SBIA & $0.835 \pm 0.055$ & $7.73 \pm 2.68 $ mm & $78.33$ \tabularnewline
\hline
Grislies & $0.834 \pm 0.082$ & $7.90 \pm 3.82 $ mm & $77.55$ \tabularnewline
\hline

\end{tabular}

\end{table}

Our implementation\footnote{Implementation available at \url{https://github.com/faustomilletari/VNet}} was realised in python, using a custom version of the Caffe\footnote{Implementation available at \url{https://github.com/faustomilletari/3D-Caffe}} \cite{jia2014caffe} framework which was enabled to perform volumetric convolutions via CuDNN v3. All the trainings and experiments were ran on a standard workstation equipped with $64$ GB of memory, an Intel(R) Core(TM) i7-5820K CPU working at 3.30GHz, and a NVidia GTX 1080 with $8$ GB of video memory. We let our model train for $48$ hours, or $30$K iterations circa, and we were able to segment a previously unseen volume in circa $1$ second. The datasets were first normalised using the N4 bias filed correction function of the ANTs framework \cite{tustison2010n4itk} and then resampled to a common resolution of $1\times1\times1.5$ mm. We applied random deformations to the scans used for training by varying the position of the control points with random quantities obtained from gaussian distribution with zero mean and $15$ voxels standard deviation. Qualitative results can be seen in Fig. \ref{fig:qualitative}.

\section{Conclusion}
We presented and approach based on a volumetric convolutional neural network that performs segmentation of MRI prostate volumes in a fast and accurate manner. We introduced a novel objective function that we optimise during training based on the Dice overlap coefficient between the predicted segmentation and the ground truth annotation. Our Dice loss layer does not need sample re-weighting when the amount of background and foreground pixels is strongly unbalanced and is indicated for binary segmentation tasks. Although we inspired our architecture to the one proposed in \cite{ronneberger2015u}, we divided it into stages that learn residuals and, as empirically observed, improve both results and convergence time. Future works will aim at segmenting volumes containing multiple regions in other modalities such as ultrasound and at higher resolutions by splitting the network over multiple GPUs. 

\section{Acknowledgement}
We would like to acknowledge NVidia corporation, that donated a Tesla K40 GPU to our group enabling this research, Dr. Geert Litjens who dedicated some of his time to evaluate our results against the ground truth of the PROMISE 2012 dataset and Ms. Iro Laina for her support to this project.

\bibliographystyle{splncs03}
\bibliography{bibliography}

\begin{thebibliography}{10}
\providecommand{\url}[1]{\texttt{#1}}
\providecommand{\urlprefix}{URL }

\bibitem{bernard2015standardized}
Bernard, O., Bosch, J., Heyde, B., Alessandrini, M., Barbosa, D., Camarasu-Pop,
  S., Cervenansky, F., Valette, S., Mirea, O., Bernier, M., et~al.:
  Standardized evaluation system for left ventricular segmentation algorithms
  in 3d echocardiography. Medical Imaging, IEEE Transactions on  (2015)

\bibitem{cha2016urinary}
Cha, K.H., Hadjiiski, L., Samala, R.K., Chan, H.P., Caoili, E.M., Cohan, R.H.:
  Urinary bladder segmentation in ct urography using deep-learning
  convolutional neural network and level sets. Medical Physics  43(4),
  1882--1896 (2016)

\bibitem{he2015deep}
He, K., Zhang, X., Ren, S., Sun, J.: Deep residual learning for image
  recognition. arXiv preprint arXiv:1512.03385  (2015)

\bibitem{huyskens2009qualitative}
Huyskens, D.P., Maingon, P., Vanuytsel, L., Remouchamps, V., Roques, T.,
  Dubray, B., Haas, B., Kunz, P., Coradi, T., B{\"u}hlman, R., et~al.: A
  qualitative and a quantitative analysis of an auto-segmentation module for
  prostate cancer. Radiotherapy and Oncology  90(3),  337--345 (2009)

\bibitem{jia2014caffe}
Jia, Y., Shelhamer, E., Donahue, J., Karayev, S., Long, J., Girshick, R.,
  Guadarrama, S., Darrell, T.: Caffe: Convolutional architecture for fast
  feature embedding. arXiv preprint arXiv:1408.5093  (2014)

\bibitem{kamnitsas2016efficient}
Kamnitsas, K., Ledig, C., Newcombe, V.F., Simpson, J.P., Kane, A.D., Menon,
  D.K., Rueckert, D., Glocker, B.: Efficient multi-scale 3d cnn with fully
  connected crf for accurate brain lesion segmentation. arXiv preprint
  arXiv:1603.05959  (2016)

\bibitem{litjens2014evaluation}
Litjens, G., Toth, R., van~de Ven, W., Hoeks, C., Kerkstra, S., van Ginneken,
  B., Vincent, G., Guillard, G., Birbeck, N., Zhang, J., et~al.: Evaluation of
  prostate segmentation algorithms for mri: the promise12 challenge. Medical
  image analysis  18(2),  359--373 (2014)

\bibitem{long2015fully}
Long, J., Shelhamer, E., Darrell, T.: Fully convolutional networks for semantic
  segmentation. In: Proceedings of the IEEE Conference on Computer Vision and
  Pattern Recognition. pp. 3431--3440 (2015)

\bibitem{milletari2016hough}
Milletari, F., Ahmadi, S.A., Kroll, C., Plate, A., Rozanski, V., Maiostre, J.,
  Levin, J., Dietrich, O., Ertl-Wagner, B., B{\"o}tzel, K., et~al.: Hough-cnn:
  Deep learning for segmentation of deep brain regions in mri and ultrasound.
  arXiv preprint arXiv:1601.07014  (2016)

\bibitem{moradi2009augmenting}
Moradi, M., Mousavi, P., Boag, A.H., Sauerbrei, E.E., Siemens, D.R.,
  Abolmaesumi, P.: Augmenting detection of prostate cancer in transrectal
  ultrasound images using svm and rf time series. Biomedical Engineering, IEEE
  Transactions on  56(9),  2214--2224 (2009)

\bibitem{noh2015learning}
Noh, H., Hong, S., Han, B.: Learning deconvolution network for semantic
  segmentation. In: Proceedings of the IEEE International Conference on
  Computer Vision. pp. 1520--1528 (2015)

\bibitem{porter2003combining}
Porter, C.R., Crawford, E.D.: Combining artificial neural networks and
  transrectal ultrasound in the diagnosis of prostate cancer. Oncology
  (Williston Park, NY)  17(10),  1395--9 (2003)

\bibitem{roehrborn1999serum}
Roehrborn, C.G., Boyle, P., Bergner, D., Gray, T., Gittelman, M., Shown, T.,
  Melman, A., Bracken, R.B., deVere White, R., Taylor, A., et~al.: Serum
  prostate-specific antigen and prostate volume predict long-term changes in
  symptoms and flow rate: results of a four-year, randomized trial comparing
  finasteride versus placebo. Urology  54(4),  662--669 (1999)

\bibitem{ronneberger2015u}
Ronneberger, O., Fischer, P., Brox, T.: U-net: Convolutional networks for
  biomedical image segmentation. In: Medical Image Computing and
  Computer-Assisted Intervention--MICCAI 2015, pp. 234--241. Springer (2015)

\bibitem{simonyan2014very}
Simonyan, K., Zisserman, A.: Very deep convolutional networks for large-scale
  image recognition. arXiv preprint arXiv:1409.1556  (2014)

\bibitem{springenberg2014striving}
Springenberg, J.T., Dosovitskiy, A., Brox, T., Riedmiller, M.: Striving for
  simplicity: The all convolutional net. arXiv preprint arXiv:1412.6806  (2014)

\bibitem{tustison2010n4itk}
Tustison, N.J., Avants, B.B., Cook, P.A., Zheng, Y., Egan, A., Yushkevich,
  P.A., Gee, J.C.: N4itk: improved n3 bias correction. Medical Imaging, IEEE
  Transactions on  29(6),  1310--1320 (2010)

\bibitem{imorp}
Vincent, G., Guillard, G., Bowes, M.: Fully automatic segmentation of the
  prostate using active appearance models. MICCAI Grand Challenge PROMISE 2012
  (2012)

\bibitem{zeiler2014visualizing}
Zeiler, M.D., Fergus, R.: Visualizing and understanding convolutional networks.
  In: Computer vision--ECCV 2014, pp. 818--833. Springer (2014)

\bibitem{zettinig2015multimodal}
Zettinig, O., Shah, A., Hennersperger, C., Eiber, M., Kroll, C., K{\"u}bler,
  H., Maurer, T., Milletari, F., Rackerseder, J., zu~Berge, C.S., et~al.:
  Multimodal image-guided prostate fusion biopsy based on automatic deformable
  registration. International journal of computer assisted radiology and
  surgery  10(12),  1997--2007 (2015)

\end{thebibliography}


\end{document}